\documentclass{article}
\PassOptionsToPackage{numbers}{natbib}
\usepackage[dblblindworkshop, final]{neurips_2026}
\workshoptitle{Neuro-Symbolic Embodied Intelligence}
\usepackage[utf8]{inputenc}
\usepackage[T1]{fontenc}
\usepackage{hyperref}
\usepackage{url}
\usepackage{booktabs}
\usepackage{amsfonts}
\usepackage{amsmath}
\usepackage{microtype}
\usepackage{xcolor}
\usepackage{colortbl}
\usepackage{xfp}
\usepackage[most]{tcolorbox}
\definecolor{PaperBlue}{HTML}{DEEAF6}
\definecolor{PaperBorder}{HTML}{B8CDD9}
\definecolor{PaperRose}{HTML}{F6C8C2}
\definecolor{PaperCream}{HTML}{FBEFD4}
\definecolor{PaperMint}{HTML}{BDE3CF}
\definecolor{PaperLilac}{HTML}{DCD3F0}
\definecolor{PaperHead}{HTML}{EAF1F9}
\newcommand{\accmix}[1]{%
  \ifdim #1pt<47.5pt
    \cellcolor{PaperCream!\fpeval{round(max(0,min(100,(#1-20)/27.5*100)),1)}!PaperRose}%
  \else
    \cellcolor{PaperMint!\fpeval{round(max(0,min(100,(#1-47.5)/27.5*100)),1)}!PaperCream}%
  \fi}
\newcommand{\acc}[1]{\accmix{#1}#1}
\newcommand{\cnt}[2]{\cellcolor{PaperLilac!\fpeval{round(max(0,min(100,#1/#2*100)),1)}!white}#1}
\newcommand{\cntb}[2]{\cellcolor{PaperLilac!\fpeval{round(max(0,min(100,#1/#2*100)),1)}!white}\textbf{#1}}
\newcommand{\hdr}{\rowcolor{PaperHead}}
\newcommand{\scorekey}{{\scriptsize Accuracy (\%):\
\colorbox{PaperRose}{20}\,%
\colorbox{PaperCream!50!PaperRose}{34}\,%
\colorbox{PaperCream}{48}\,%
\colorbox{PaperMint!50!PaperCream}{61}\,%
\colorbox{PaperMint}{75}}}
\newtcolorbox{promptbox}[1]{
    enhanced,colback=PaperBlue!25!white,colframe=PaperBorder,
    colbacktitle=PaperBlue,coltitle=black,fonttitle=\small\bfseries,
    fontupper=\small\ttfamily,before upper=\raggedright,
    title={#1},boxrule=0.5pt,arc=2mm,left=9pt,right=9pt,top=7pt,bottom=7pt,
    before skip=8pt,after skip=8pt
}
\usepackage{multirow}
\usepackage{graphicx}
\usepackage{tikz}
\usetikzlibrary{arrows.meta,positioning}
\usepackage{placeins}
\usepackage[capitalize]{cleveref}
\hypersetup{hidelinks}
\newcommand{\rev}[1]{#1}
\newcommand{\accr}[1]{\accmix{#1}\rev{#1}}

\title{Answering Without Arriving:\\
An Evidence Audit for Embodied Question Answering}
\author{%
  Shubh Garg$^{1}$\thanks{Equal contribution, co-first authors.}\quad
  Divyansh Pandey$^{2}$\footnotemark[1]\quad
  Niyati Rawal$^{1,3}$\quad
  Vedant Khandelwal$^{4}$\quad
  Amit Sheth$^{1}$ \\[2pt]
  \normalfont $^{1}$Indian AI Research Organisation (IAIRO), GIFT City, Gujarat\quad
  $^{2}$Independent Researcher \\
  \normalfont $^{3}$BITS Pilani\quad
  $^{4}$AI Institute, UofSC (AIISC), 1112 Greene St., Columbia \\[2pt]
  \normalfont\small\texttt{\{shubh.garg@iairo.ai, shubhgarg265@gmail.com\}, divyanshpandey0108@gmail.com,} \\
  \normalfont\small\texttt{niyati.rawal@iairo.ai, kh.vedant02@gmail.com,} \\
  \normalfont\small\texttt{\{amit.sheth@iairo.ai, amitpsheth@gmail.com\}}
}
\begin{document}
\maketitle

\begin{abstract}
Embodied question answering asks an agent to navigate, observe, and then
answer questions about what it saw, yet answer accuracy alone cannot tell
whether any of this happened. A plausible answer can come from the question,
the room label, or the answer choices, and a correct answer does not show
that the supporting observation was ever made. We present a multi-room
benchmark whose 200 test episodes and 800 questions are stratified by route
extent, with an evaluation ladder that holds the questions fixed while
varying only the evidence supplied to the answerer, from text-only probes
through closed-loop agents to replay of the reference trajectory. Questions
are generated from symbolic records of rooms, visible objects, and frames,
so every answer can be traced to the observations that support it. A
closed-loop vision-language agent reaches no target in 200 episodes yet
answers 45.5\% of questions, within 0.1 points of the same model given no
images; replaying the reference trajectory raises the same model to 57.4\%,
so the evidence exists along the route but is not gathered. A family-level
audit then shows that access to evidence is not the same as use of it. Four
of eight families are largely answerable from text, one cites frames that
cannot establish its own answer, and the oracle conditions reach 59.5\% on
Rooms-visited by counting the supplied images. Evidence acquisition, answer
coverage, and shortcut sensitivity must therefore be established before an
accuracy gain can be credited to memory.
\end{abstract}

\section{Introduction}
\label{sec:intro}

A household robot sent to find the sink passes a kettle, a sofa, and a
houseplant on the way. Asked afterwards which room held the kettle, it needs
no new perception, because the evidence was already in view. What it needs is
to have reached a useful viewpoint, retained the observation, and retrieved it
when asked. Embodied question answering (EQA) couples these steps, but the
accuracy score that summarises an EQA run cannot say which of them occurred.
A correct answer may come from the question wording, the room label, or the
answer choices rather than from anything the agent saw
\citep{anand2018,gururangan2018}. An incorrect answer may mean that the agent
never observed the evidence, that it observed and lost it, or that the
observations handed to the answerer did not support the question.

This ambiguity is acute for neuro-symbolic embodied systems, in which
structured memories organise observations into entities, relations, and
temporal records that a planner and an answerer can query
\citep{saxena2024grapheqa,graphpad2025,stkg2025}. Crediting an accuracy gain
to such a memory requires that the observations, the symbolic records, and
the questions all refer to the same events, so that one can check what was
observed and whether the answer could have been produced without it. The
generator that writes the questions already holds this information, as room
labels, visible-object logs, and frame indices; the evaluation should use it.

We present a multi-room EQA benchmark and evaluation ladder built on this
principle. Episodes traverse procedurally generated ProcTHOR houses along a
reference route whose extent is fixed at generation time, from one to four or
more rooms. \rev{Symbolic templates write multiple-choice questions from the
simulator's visible-object log over a room-labelled strip of trajectory
frames, and each question is stored with the frame indices that support it.}
The ladder keeps the 800 test questions and their format
fixed while varying only the evidence available to the answerer, from
text-only probes and closed-loop agents that gather their own observations
to oracle conditions that cite or replay the reference trajectory.

The ladder yields a headline that accuracy alone would conceal. A closed-loop
VLM agent emits a valid action at every step, reaches no target in 200
episodes, and still answers 45.5\% of questions, within 0.1 points of the
same model given no images. Replaying the reference trajectory raises the
same model to 57.4\%, so the evidence the questions need is present along the
route and is simply not gathered. A family-level audit then shows that access
to evidence is not the same as use of it. Four of the eight question families
are largely answerable from text, the cited frames of one family can omit the
observation that establishes its answer, and the oracle conditions can score
59.5\% on Rooms-visited by counting the images they receive.

Our contributions are threefold:
\begin{enumerate}
    \setlength{\itemsep}{0pt}
    \item A multi-room EQA benchmark with controllable route extent,
    trajectory-linked question records, and explicit symbolic checks of
    target leakage and answer support (\cref{sec:benchmark}).
    \item An evaluation ladder that reports navigation, answer coverage, and
    per-family accuracy separately, exposing a closed-loop agent whose
    answer accuracy conceals complete navigation failure (\cref{sec:ladder}).
    \item A validity audit that identifies text-prior sensitivity,
    evidence-selection defects, and an input-count shortcut, and derives the
    controls a memory comparison must satisfy (\cref{sec:validity}).
\end{enumerate}
Together these give a measurement framework in which evidence acquisition,
retention, and answering can be tested separately, a precondition
for evaluating structured memory in embodied agents.

\section{Related work}
\label{sec:related}

\textbf{EQA and evidence-free baselines.}
EmbodiedQA couples navigation with answering questions about an environment
\citep{das2018}. Blindfold baselines showed that question-only models exploit
regularities in EQA data \citep{anand2018}, echoing annotation artifacts in
language-inference datasets \citep{gururangan2018}. OpenEQA distinguishes
episodic-memory and active-exploration settings and reports blind,
frame-based, and scene-graph baselines \citep{majumdar2024}. Our text-only
controls follow this tradition, with one difference that matters for
interpretation. Our blind condition also receives the target-room label, so
it is a room-conditioned control rather than a reproduction of OpenEQA's
question-only baseline.

\textbf{Exploration and memory benchmarks.}
EXPRESS-Bench records exploration trajectories and scores the consistency
between observations and answers \citep{jiang2025}; its questions are
generated from the terminal observation, whereas ours also refer to events
earlier along the reference route. Memory-centric EQA introduces multi-target
questions in MT-HM3D \citep{memcentric2025}, and FindingDory evaluates memory
over extended embodied tasks \citep{findingdory2025}. We add two elements
that these benchmarks do not combine, namely route extent controlled at
generation time and a same-model comparison between closed-loop and replayed
evidence that is read jointly with a per-family text-prior audit.

\textbf{Structured scene memory.}
GraphEQA grounds planning in online 3D scene graphs and task-relevant images
\citep{saxena2024grapheqa}; GraphPad edits a scene graph at inference time
through a scratchpad interface \citep{graphpad2025}; 3D-Mem stores explored
and frontier regions as memory snapshots \citep{mem3d2024}; and
spatiotemporal knowledge graphs can be built from monocular video without 3D
reconstruction \citep{stkg2025}. MomaGraph trains a VLM to predict
state-aware unified scene graphs for embodied task planning
\citep{ju2026momagraph}. These systems motivate the question the benchmark is
built to ask, whether a structured memory preserves the evidence an answer
needs. We include the released MomaGraph-R1 model as a specialised baseline,
both as a scene-graph-guided closed-loop policy and as a frame-oracle
answerer, and the audit prepares the matched memory ablation that would
answer the question directly.

\section{Benchmark and evaluation design}
\label{sec:benchmark}

\subsection{Task and reference trajectories}

\begin{figure}[t]
      \centering
      \includegraphics[width=\linewidth]{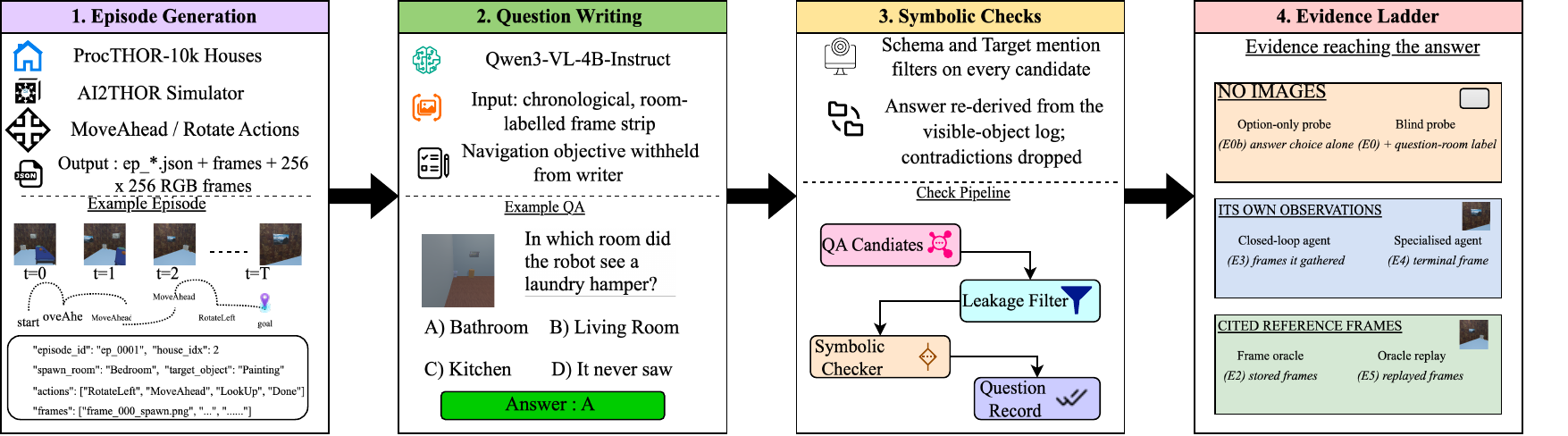}
       \caption{Benchmark and evaluation pipeline. (A) Episode generation produces a                               
  reference route and a privileged symbolic record; the example strip shows a bed
  visible at $t{=}0$ and $t{=}1$ and gone by $t{=}2$. (B) \rev{Symbolic
  templates write} multiple-choice questions from that strip, with the
  navigation objective withheld. The question shown is answerable from the full trajectory, but its
  cited frames $\{0,1\}$ omit the later absence, the evidence-selection defect of
  \Cref{sec:validity}. (C) Schema, target-mention, and symbolic checks link each
  question to the frames supporting it. (D) The ladder holds the 800 questions and
  their format fixed and varies only the evidence reaching the answerer, which
  never includes the privileged record.}
      \label{fig:architecture}
  \end{figure}

An episode consists of a house, spawn pose, navigation target instance, reference
trajectory, and associated questions. The agent navigates toward the target
and then answers questions about the reference trajectory. The navigation
objective is therefore distinct from the incidental observations queried
afterwards. When a policy follows a different route, it may never observe
the events on which a question's gold answer depends. We retain those
questions because the resulting score is the end-to-end outcome a deployed
system would produce, while treating access to their evidence as a necessary
control for any memory-specific conclusion. \Cref{fig:architecture}
summarises the pipeline.

Episodes use ProcTHOR houses \citep{deitke2022} in AI2-THOR 5.0.0
\citep{kolve2017}, with headless \texttt{CloudRendering} and
$256\times256$ RGB observations. A room graph specifies a spawn room and
target room connected by a chosen sequence of doorways. A hierarchical
planner converts this route into a reference action sequence. The task
involves navigation and observation, without object manipulation. Object
visibility refers to the camera view; disappearance means leaving that view.
\rev{An object, including a partly occluded or cut-off one, enters a frame's
visible-object record when AI2-THOR flags it visible and its visible region
covers at least 0.0005 of the image (about 33 pixels at $256\times256$).}

\textbf{Action space and generation checks.}
The generator records parameterised actions: \texttt{MoveAhead} at
$\{0.25,0.5,1.0\}$\,m; left/right rotations at
$\{15,30,45,90\}^{\circ}$; look actions at $\{15,30\}^{\circ}$ within
$[-30,60]^{\circ}$ pitch; and \texttt{Done}. Each episode records the action vocabulary and magnitudes needed for
replay. Reference routes use fully discrete execution with zero doorway
teleport hops; the simulator restores the recorded spawn pose before execution.

The target must be invisible at spawn. Minimum spawn distance is
$3.0\times$ the 1.5\,m stop distance for cross-room episodes and
$1.5\times$ for same-room episodes. Generation accepts a trajectory after
walking it to the target under the discrete action menu. Success requires the
specific \texttt{target\_objectId} to lie within 1.5\,m and occupy at least
0.005 of the image area. This instance-based criterion ties success to the designated object.
The navigation grid is 0.25\,m, visibility extends to 6.0\,m, and generation
uses a 400-step cap. Evaluation allocates three times the reference action
length to each episode. \Cref{tab:constants} collects these controls.

\subsection{Questions and symbolic validation}

\rev{Symbolic templates read the simulator's per-frame visible-object log over
a chronological, room-labelled strip of up to six trajectory frames and
produce each question, its reference answer, and a justification. The
navigation objective is withheld, and each question has two to four shuffled
options. Because answers are derived from the symbolic record, they agree with
it by construction; questions also pass schema checks and a check for explicit
target mentions.} Each question stores
its family, choices, answer, frame references, and a memory-requirement flag,
linking the language output to structured episode evidence. The subsequent
audit evaluates text priors and the evidence supplied to each answerer.

The eight families are Existence, Counting, Room Identification, Order of
first observation, Persistence across rooms, Presence-change, Global-count,
and Rooms-visited. Global-count and Rooms-visited query trajectory-level quantities. Family names describe
the intended query, not a guarantee that every instance requires memory.
For example, a whole-route count may be recoverable from a short route, and
an evidence reference can inadvertently reveal the answer through its length.
\Cref{sec:validity} evaluates these limitations before interpreting accuracy.

\textbf{Dataset scope.}
The benchmark comprises 5,000 training, 500 validation, and 200 test episodes
drawn from 1,359, 151, and 53 houses, respectively. The designated test set
contains 800 questions. Test episodes are balanced across S1--S4, with 50
episodes for each of 1, 2, 3, and at least 4 distinct rooms on the reference
route. Houses are drawn from ProcTHOR-10k's respective train, validation, and
test partitions. The frame cache covers all 206,899 training, 21,982 validation,
and 8,335 test observations. Split statistics and
house-identity conventions appear in \cref{app:stats,app:splits}.

\subsection{Evidence conditions}
\label{sec:protocol}

The evaluation ladder assigns the same designated questions to several
conditions. Each uses the same answer choices, but actual scoring coverage
can differ when execution or answering fails.
\begin{itemize}
    \setlength{\itemsep}{0pt}
    \item \textbf{E0b, options-only:} answer choices without the question,
    images, or trajectory.
    \item \textbf{E0, blind:} question, choices, and target-room label, with
    no images. This measures room-conditioned language-prior performance.
    \item \textbf{E1, trivial navigation:} immediate-\texttt{Done} and
    random-action policies, without a model-generated answer channel.
    \item \textbf{E2, frame oracle:} question and choices with the stored
    frames cited by that question. Navigation and frame selection are supplied.
    \item \textbf{E3, closed loop:} a VLM chooses an action from the current
    image, steps the simulator, and repeats. After termination it answers
    from a subsample of its own observations, not the reference frames.
    \item \textbf{E4, specialised closed loop:} MomaGraph-R1 predicts a local
    scene graph, selects an action, and answers from the terminal image after
    emitting \texttt{Done} (\cref{app:e4}).
    \item \textbf{E5, oracle replay:} the simulator executes the reference
    actions, and the answerer receives the replayed images indexed by each
    question's frame references.
\end{itemize}
\rev{E0b, E0, E2, E3, and E5 all run on Qwen3-VL-4B-Instruct, so the main
ladder holds the model fixed. E4 uses MomaGraph-R1, and MomaGraph-R1 and
Qwen2.5-VL-7B-Instruct give two cross-model E2 references.} E3 and E5 share
a model, but change both trajectory execution and evidence
selection. E2 and E5 measure answering with privileged trajectory evidence.
The text-only runner passes an empty image list regardless of stored frame
references, enforcing the blind control.

\subsection{Metrics and execution controls}

Navigation success rate (SR) uses the target-instance criterion above.
Success weighted by path length (SPL) is
\begin{equation}
    \mathrm{SPL}=\frac{1}{N}\sum_{i=1}^{N}
    S_i\frac{\ell_i}{\max(\ell_i,p_i)},
\end{equation}
where $S_i$ indicates success, $\ell_i$ is the reference shortest-path
distance, and $p_i$ is the executed path length \citep{anderson2018spl}.
We also retain distance reduction
$d_\Delta=d_{\mathrm{init}}-d_{\mathrm{final}}$ and executed path length,
which describe progress and movement even when a target is not reached.

Action-format compliance records whether an emitted string maps to an allowed
action name. It is distinct from whether the simulator executes that action
successfully. Episodes terminate on \texttt{Done}, budget exhaustion, or a
stuck detector after eight consecutive steps without positional progress.
Rotating or looking in place therefore counts toward the stuck threshold,
even when it collects new visual information, so reported outcomes describe
the policy and harness together.

For each question, the answer channel records the raw response, parsed
letter, and parse-success flag. We report answer accuracy alongside parse coverage and distinguish
accuracy over all questions from accuracy on parsed responses. All model conditions use greedy decoding
and a short answer budget. An A/B/C/D instruction is retained even when
fewer choices exist. Chance is the mean of $1/K_q$ over the $K_q$ choices
of each evaluated question, giving 33.02\% for the test option-count
summary (\cref{app:chance}). Shaded cells map answer accuracy from rose
(20\%) through cream to mint (75\%) on one shared display scale; the printed
values are exact.

\subsection{Implementation and compute}
\label{sec:hardware}
Each condition runs on one NVIDIA H200 GPU without 4-bit quantisation or
model parallelism. The generic CUDA runner uses \texttt{bfloat16}; the E4
adapter uses \texttt{float16}. The execution node has eight H200
GPUs (143,771\,MiB each), two AMD EPYC 9575F 64-core processors, and
2,267\,GB system RAM. The software stack is Python 3.12.13, PyTorch 2.11.0
with CUDA 13.0, Transformers 5.7.0, and AI2-THOR 5.0.0. Headless
\texttt{CloudRendering} uses Vulkan (models: \cref{tab:models}).

The E3 runner parses a base action name and applies the episode's
configured movement or camera increment. It supplies the current RGB view
for action selection and uniformly subsamples its captured views for final
QA. Oracle replay instead executes the recorded parameterised actions and
resolves each question's frame references in the replayed sequence. Action selection is reactive, receiving only the current view and
target type, with no persistent action history or graph memory. The prompt
requests \texttt{Done} when that type is visible; the scorer separately
checks the target instance, distance, and image area
(\cref{app:prompts,app:examples}).

\begin{table}[t]
    \centering
    \footnotesize
    \caption{Generation and execution settings. Values are recorded per episode in the
    released data.}
    \label{tab:constants}
    \begin{tabular}{lll}
        \toprule
        \hdr Constant & Value & Effect \\
        \midrule
        \texttt{stop\_dist} & 1.5\,m & instance-based success radius \\
        \texttt{min\_spawn\_dist\_mult} & $3.0\times$ (4.5\,m) & minimum spawn--target distance \\
        \texttt{min\_spawn\_dist\_mult\_same\_room} & $1.5\times$ (2.25\,m) & same-room episodes \\
        \texttt{visibility\_distance} & 6.0\,m & simulator visibility cutoff \\
        \texttt{min\_bbox\_area\_frac} & 0.005 & minimum target frame coverage \\
        \rev{\texttt{perception\_min\_bbox\_frac}} & \rev{0.0005} & \rev{minimum area to log an object as visible} \\
        \texttt{eval\_step\_budget\_k} & $3\times$ GT length & per-episode step budget \\
        \texttt{max\_nav\_steps} & 400 & generation-time hard cap \\
        \texttt{grid\_size} & 0.25\,m & navmesh grid \\
        \texttt{move\_menu\_m} & $\{0.25, 0.5, 1.0\}$ & \texttt{MoveAhead} magnitudes \\
        \texttt{rotate\_menu\_deg} & $\{15, 30, 45, 90\}$ & rotation angles \\
        \texttt{horizon\_menu\_deg} & $\{15, 30\}$ & look angles \\
        \texttt{frame\_size} & $256\times256$ & RGB observation \\
        Stuck criterion & 8 steps & consecutive steps without positional progress \\
        \texttt{teleport\_hops} & 0 & every episode fully discrete \\
        \bottomrule
    \end{tabular}
\end{table}

\begin{table}[t]
    \centering
    \footnotesize
    \caption{Model revisions. Inference uses greedy decoding and unquantised
    16-bit weights.}
    \label{tab:models}
    \begin{tabular}{llll}
        \toprule
        \hdr Model & Revision & Used for & Params \\
        \midrule
        \texttt{Qwen/Qwen3-VL-4B-Instruct} & \texttt{ebb281ec} & E0b, E0, \rev{E2,} E3, E5 & 4B \\
        \texttt{cheryyunl/MomaGraph-R1} & \texttt{d447423d} & E2, E4 & 8B \\
        \texttt{Qwen/Qwen2.5-VL-7B-Instruct} & \texttt{cc594898} & E2 & 7B \\
        \bottomrule
    \end{tabular}
\end{table}

\section{Validity audit}
\label{sec:validity}

\textbf{Text-only performance.}
Options-only reaches 39.0\% and blind reaches 45.4\%, with 800 and 796 parsed
responses, respectively, out of 800 questions. Both exceed the pooled 33.02\%
chance reference, establishing a substantial evidence-free component in
aggregate accuracy.

\begin{table}[t]
    \centering
    \footnotesize
    \caption{Text-only probe accuracy by question family. Probe-sensitive
    families show strong performance without images. The screening describes
    observed probe behaviour and evidence requirements.}
    \label{tab:validity}
    \begin{tabular}{lcccl}
        \toprule
        \hdr Family & $n$ & Opt.-only (\%) & Blind (\%) & Screening \\
        \midrule
        Persistence         & 131 & \acc{51.9} & \acc{61.1} & Probe-sensitive \\
        Global-count        & 120 & \acc{31.7} & \acc{33.3} & Low probe scores \\
        Rooms-visited       & 116 & \acc{32.8} & \acc{28.4} & Low probe scores \\
        Existence           & 113 & \acc{31.0} & \acc{57.5} & Probe-sensitive \\
        Presence-change     & 102 & \acc{49.0} & \acc{36.3} & Evidence mismatch \\
        Room Identification & 80  & \acc{28.7} & \acc{65.0} & Probe-sensitive \\
        Order               & 73  & \acc{60.3} & \acc{43.8} & Probe-sensitive \\
        Counting            & 65  & \acc{24.6} & \acc{36.9} & Moderate blind score \\
        \bottomrule
    \end{tabular}
\end{table}

\textbf{Family-level screening.}
Persistence, Existence, Room Identification, and Order collectively contain
397 of the 800 questions and show strong text-only performance
(\cref{tab:validity}). The options-only result for Order (60.3\%) makes
answer-choice regularity a particular concern. Room Identification is
different, because its 65.0\% blind score is measured with a target-room
label in the prompt, so a room-label ablation is needed to separate that cue
from other language priors.

Rooms-visited and Global-count have lower text-only scores, making them
useful candidates for evidence-controlled evaluation; low probe accuracy
alone does not prove memory dependence.

\textbf{Evidence selection in Presence-change.}
The generator can ask whether an object disappeared by the end of the route
while citing only its first and last visible frames. Those frames show the
object present and omit the later observation needed to establish absence.
This is an evidence-selection defect even when the gold answer is correct for
the full trajectory. The family scores 36.3\% blind, 33.3\% in E2, and
39.6\% in E5, compared with 61.8\% in E3. This contrast makes the distinction between full-trajectory answer support
and cited-frame sufficiency central to interpreting this family. \Cref{app:presence_change,app:examples} traces the mechanism to
code and a worked sample.

\textbf{Input-count leakage in the oracles.}
E2 and E5 supply the frames cited by each question. On Rooms-visited, the
number of cited frames equals the gold room count in 69 of 116 questions. A rule that answers with that number therefore scores 59.5\%,
without inspecting image content, compared with a 39.7\% majority-answer
baseline. The oracle scores of 67.2\% and 68.1\% exceed this simple rule,
but their margins are not an estimate of the causal contribution of vision.
A model may combine input count, answer choices, and other cues. Establishing
visual evidence use requires a matched experiment that holds image count
fixed or removes image content (\cref{app:shortcut}).

\section{Results}
\label{sec:ladder}

\begin{table}[t]
    \centering
    \footnotesize
    \caption{Per-family answer accuracy (\%). $n$ gives the test-set
    family size. E0b, E0, E3, and E5 use Qwen3-VL-4B; E2 uses MomaGraph-R1
    \rev{(MG) or Qwen3-VL-4B (Qw)}.
    E5 accuracy uses its scored questions. E4 answer coverage is detailed in
    \cref{app:e4}.}
    \label{tab:perfamily}
    \scorekey\par\smallskip
    \begin{tabular}{lrrrrrrr}
        \toprule
        \hdr Family & $n$ & E0b & E0 & E3 & E2 \rev{(MG)} & \rev{E2 (Qw)} & E5 \\
        \midrule
        Persistence         & 131 & \acc{51.9} & \acc{61.1} & \acc{61.1} & \acc{66.4} & \accr{68.7} & \acc{69.5} \\
        Global-count        & 120 & \acc{31.7} & \acc{33.3} & \acc{27.5} & \acc{32.5} & \accr{30.0} & \acc{30.5} \\
        Rooms-visited       & 116 & \acc{32.8} & \acc{28.4} & \acc{32.8} & \acc{67.2} & \accr{67.2} & \acc{68.1} \\
        Existence           & 113 & \acc{31.0} & \acc{57.5} & \acc{40.7} & \acc{68.1} & \accr{73.5} & \acc{71.4} \\
        Presence-change     & 102 & \acc{49.0} & \acc{36.3} & \acc{61.8} & \acc{33.3} & \accr{40.2} & \acc{39.6} \\
        Room Identification & 80  & \acc{28.7} & \acc{65.0} & \acc{46.3} & \acc{68.8} & \accr{66.2} & \acc{65.4} \\
        Order               & 73  & \acc{60.3} & \acc{43.8} & \acc{57.5} & \acc{60.3} & \accr{74.0} & \acc{73.2} \\
        Counting            & 65  & \acc{24.6} & \acc{36.9} & \acc{38.5} & \acc{43.1} & \accr{50.8} & \acc{46.2} \\
        \midrule
        Overall             & 800 & \acc{39.0} & \acc{45.4} & \acc{45.5} & \acc{55.2} & \accr{58.5} & \acc{57.4} \\
        \bottomrule
    \end{tabular}
\end{table}

\subsection{Answer accuracy can conceal failed navigation}

E3 reaches no target in 200 episodes, although its action-format
compliance is 100\% and its invalid-action rate is 0\%. It executes a mean of
12.0 actions (median 9), travels 1.03\,m, and terminates as stuck in 172
episodes. Its 45.5\% answer accuracy is numerically close to E0's 45.4\%.
Together, these metrics show that fluent action selection and correct
answers can coexist with limited evidence acquisition.

The trivial E1 policies also reach no target and have no answer channel.

E4 also reaches no target, travels 0.46\,m on average, and ends as stuck in
168 episodes. Only 128 of 800 questions have parsed answers; 54 of those are
correct (42.2\%). This is a conditional score on a selected subset, with an
unconditional accuracy of 6.8\%. The shared simulator and scorer compare a reactive VLM
policy with a scene-graph-guided policy with different answer triggers (\cref{app:e4}).

\subsection{Answering with reference-trajectory evidence}

E5 reports 57.4\% accuracy, a difference of 11.9 percentage points from E3.
This same-model comparison measures the advantage of the replay evidence
protocol, which supplies both a reference route and question-specific frame
selection. Oracle replay reaches every replay-scored target (SPL 0.924).

\rev{With Qwen3-VL-4B, E2 reaches 58.5\%, within 1.1 points of E5 and 13.0
points above the same model's closed-loop run, so the gap between gathered
and supplied evidence does not rest on a change of model.} E2 reports 55.2\%
with MomaGraph-R1 and 55.0\% with Qwen2.5-VL-7B. MomaGraph-R1's single-frame
QA path scores 44.1\%, below its 55.2\% multi-frame condition, a comparison
that varies the QA configuration as well as the available frames.

\begin{figure}[t]
    \centering
    \includegraphics[width=0.94\linewidth]{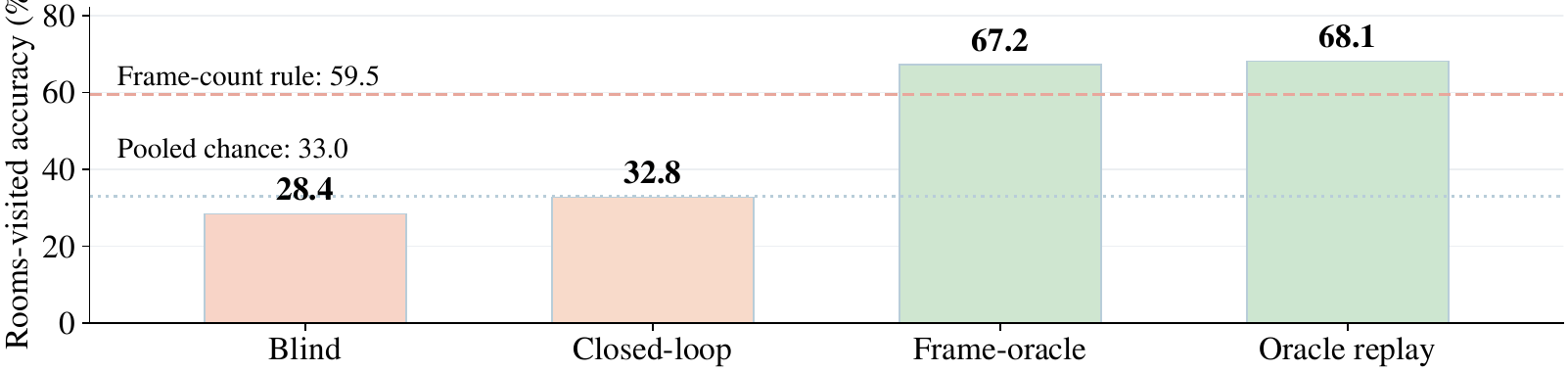}
    \caption{Rooms-visited accuracy ($n=116$). The horizontal
    33.0\% line is the pooled test-set chance reference. The 59.5\% line is a content-free
    rule that answers with the number of supplied images. Oracle scores
    exceed this rule, but the excess does not isolate visual reasoning.
    E0, E3, and E5 use Qwen3-VL-4B; E2 uses MomaGraph-R1.}
    \label{fig:rooms_visited}
\end{figure}

\subsection{Family and route-level differences}

Rooms-visited has blind accuracy 28.4\% and closed-loop accuracy 32.8\%,
both below the 39.7\% majority-answer baseline. Oracle conditions
reach 67.2\% and 68.1\%. This contrast makes the family useful for studying
evidence access, subject to controlling image-count leakage. It uses a single
question template, and only 63 of its 116 instances carry the generator's
\texttt{memory\_required} flag. Global-count scores 32.5\% and 30.5\% in the two oracles, identifying
trajectory-wide counting as a separate test of evidence coverage and
aggregation.

\begin{table}[t]
    \centering
    \footnotesize
    \caption{Accuracy and navigation by route-length stratum
    (50 designated test episodes each). $\Delta$ subtracts the
    stratum-specific chance value (0.322--0.334). Navigation entries give
    the successful and scored episodes for each condition.}
    \label{tab:strata}
    \begin{tabular}{llrrrr}
        \toprule
        \hdr Rung & Metric & S1 & S2 & S3 & S4 \\
        \midrule
        \multirow{2}{*}{E3 (closed-loop)} & EQA (\%) & \acc{49.5} & \acc{44.5} & \acc{50.0} & \acc{38.0} \\
         & $\Delta$ vs chance & $+16.2$ & $+11.1$ & $+17.8$ & $+4.8$ \\
        \midrule
        \multirow{2}{*}{E5 (oracle replay)} & EQA (\%) & \acc{61.5} & \acc{52.5} & \acc{60.0} & \acc{55.5} \\
         & $\Delta$ vs chance & $+28.2$ & $+19.1$ & $+27.8$ & $+22.3$ \\
        \midrule
        \multicolumn{2}{l}{E5 $-$ E3 (points)} & $+12.0$ & $+8.0$ & $+10.0$ & $+17.5$ \\
        \midrule
        \multirow{3}{*}{E3 navigation} & SR & 0/50 & 0/50 & 0/50 & 0/50 \\
         & path (m) & 0.71 & 1.16 & 0.77 & 1.49 \\
         & stuck & 42 & 45 & 46 & 39 \\
        \midrule
        \multirow{3}{*}{E5 navigation} & SR & 50/50 & 49/49 & 50/50 & 49/49 \\
         & SPL & 0.966 & 0.940 & 0.917 & 0.875 \\
         & path (m) & 6.58 & 11.32 & 16.10 & 22.91 \\
        \bottomrule
    \end{tabular}
\end{table}

The E5--E3 accuracy difference is largest in S4 (17.5 points), but
varies non-monotonically across S1--S4 (12.0, 8.0, 10.0, 17.5 points).
Because the strata also differ in question composition, we read these gaps
as route-stratified performance, not a pure route-length effect.

\section{Discussion}
\label{sec:limitations}

\textbf{Implications for neuro-symbolic evaluation.}
The benchmark ties \rev{trajectory questions} to symbolic room, object, action,
and observation records, and the audit shows why each interface between them
needs its own check. A symbolic re-derivation can certify a gold label while
the cited frames fail to support it (Presence-change), and a structurally
valid evidence set can leak the answer through its size (Rooms-visited). Crediting a graph memory requires evidence that
was observed, retained, and used; the ladder tests the first.

\rev{\textbf{Why oracle accuracy stays below the symbolic record.}
The symbolic record answers every question by construction, while E5 reaches
57.4\%. Logged objects can be as small as 33 pixels, cited frames can omit
the deciding observation (Presence-change), and Global-count aggregates
across frames while its wording refers to the house but its reference counts
only the sampled strip. Higher resolution, evidence selection that includes
the deciding frame, and trained readers of a structured record are direct
remedies. Replay is only an evaluation oracle; E2 reuses stored frames
(58.5\%).}

\rev{\textbf{What the ladder does and does not diagnose.}
The ladder separates evidence acquisition (E3 versus E5) from answering
with supplied evidence (E2, E5). It does not separate retention from
retrieval or integration; that needs textual, vector, key--value, and
scene-graph memories read on the same observed trajectory. A companion study on this benchmark trains
answerers to read one scene-graph record as text, learned graph features, or
both. This study is diagnostic only.}

\textbf{Evaluation scope.}
The study evaluates automatically generated questions in ProcTHOR/AI2-THOR
using one run per condition. \rev{Rendering is simpler than in photorealistic scans
such as those used by OpenEQA \citep{majumdar2024}; only the visual rungs
depend on it.} The eight-step
positional stopping rule also makes observation collection sensitive to
in-place camera actions.

\textbf{Next experimental controls.}
Priority controls are to include the final absence observation, balance and
re-probe shortcut-sensitive families, fix the oracle image budget independently
of each answer, and compare memory variants on identical successfully
observed trajectories. A content-ablated oracle and a room-label ablation
would test two cues exposed here. \rev{The label can also mislead:
some blind responses treat it as the robot's location and decline to answer
about another room. Renaming or shuffling room labels, or placing objects in
atypical rooms, would measure this reliance.}

\section{Conclusion}
We present a trajectory-based embodied QA benchmark that connects controlled
navigation routes, symbolic observation records, and an evidence-access
ladder. The closed-loop agent scores 45.5\% without reaching a
single target, while the blind control scores 45.4\%. Replaying the
reference trajectory raises the same model to 57.4\%: the missing accuracy
lies in evidence never gathered, not in the questions. Image access alone
does not establish image use; text priors, frame sufficiency, and input-count
cues all affect that reading.

\begin{ack}
The authors received no funding for this work and declare no competing interests.
\end{ack}

\bibliographystyle{plainnat}
\bibliography{refs}
\newpage
\appendix

\section{Benchmark statistics}
\label{app:stats}
\Cref{tab:benchmark_stats} gives split sizes, house counts, and complete
frame-cache coverage. Multiple episodes can share a house with different
spawn poses and navigation targets. House identity therefore defines the
unit of split separation, while the question family defines the grouping
used for answer analysis.

\begin{table}[h]
    \centering
    \small
    \caption{Benchmark statistics. Question counts are for the 800-question test
    split; the per-family probe screening derived from these questions is
    in \cref{tab:validity}.}
    \label{tab:benchmark_stats}
    \begin{tabular}{lccc}
        \toprule
        \hdr & Train & Val & Test \\
        \midrule
        Episodes & 5{,}000 & 500 & 200 \\
        Distinct houses & 1{,}359 & 151 & 53 \\
        Frame cache coverage & 206{,}899/206{,}899 & 21{,}982/21{,}982 & 8{,}335/8{,}335 \\
        \bottomrule
    \end{tabular}
    \hfill
    \begin{tabular}{lc}
        \toprule
        \hdr Test questions per family & $n$ \\
        \midrule
        Persistence & \cnt{131}{131} \\
        Global-count & \cnt{120}{131} \\
        Rooms-visited & \cnt{116}{131} \\
        Existence & \cnt{113}{131} \\
        Presence-change & \cnt{102}{131} \\
        Room Identification & \cnt{80}{131} \\
        Order & \cnt{73}{131} \\
        Counting & \cnt{65}{131} \\
        \midrule
        Total / mean chance & 800 / 33.0\% \\
        \bottomrule
    \end{tabular}
\end{table}

\section{Chance computation}
\label{app:chance}
For question set $\mathcal Q$, uniform-guess accuracy is
\begin{equation}
    c(\mathcal Q)=\frac{1}{|\mathcal Q|}
    \sum_{q\in\mathcal Q}\frac{1}{K_q}.
\end{equation}
The test set has 175 two-option, 245 three-option, and 380
four-option questions. Its pooled chance is
$(175/2+245/3+380/4)/800=33.02\%$. The stratum references range from
0.322 to 0.334. A pooled chance value is not interchangeable with a
family-specific reference or a majority-answer baseline. The same formula applies to any family or stratum by restricting
$\mathcal Q$ to its questions.

\section{House identities and split definitions}
\label{app:splits}
The construction assigns houses from ProcTHOR-10k's respective training,
validation, and test partitions. Episode files store within-split house
indices, so the pair (partition, index) identifies a house. Reusing an index
across partitions refers to separate source records. Split-aware identities
or house-content hashes support comparisons across exported datasets.

\section{Prompt and evidence interfaces}
\label{app:prompts}
The text-mode runner does not resolve image references. Options-only receives
only choices, whereas blind receives the question and target-room label.
The following boxes give the templates.
\begin{promptbox}{Options-only prompt}
Options:\\
A: Before\\
B: After\\[3pt]
Pick the most plausible answer. Answer with a single letter (A/B/C/D).
\end{promptbox}
\begin{promptbox}{Blind prompt}
Room: Bathroom\\[3pt]
Question: Did the robot first see a sink before or after it first saw a
houseplant?\\
Options:\\
A: Before\\
B: After\\[3pt]
Answer with a single letter (A/B/C/D).
\end{promptbox}
The oracle template is:
\begin{promptbox}{Frame-oracle prompt}
Question: \{question\}\\
Options:\\
\{options\}\\[3pt]
Look at the image(s). Answer with a single letter (A/B/C/D).
\end{promptbox}
\subsection{E3 action and answer prompts}
The action prompt names the target type and accompanies only the current
RGB frame. The instance identifier remains part of the scorer's target
record. The implementation uses this template:
\begin{promptbox}{E3 navigation prompt}
You are a robot exploring a house, looking for a \{target\_type\}.
Here is what you currently see. Choose exactly one action to move
closer to finding it, or to confirm you can see it now:\\
MoveAhead, RotateLeft, RotateRight, LookUp, LookDown, Done.\\
Say Done only if the \{target\_type\} is visible in the current image.\\
Respond with exactly one action name.
\end{promptbox}
After termination, the answer prompt accompanies an evenly spaced selection
of the agent's captured frames:
\begin{promptbox}{E3 answer prompt}
Question: \{question\}\\
Options:\\
\{options\}\\[3pt]
Based on everything you saw while exploring, answer with a single
letter (A/B/C/D).
\end{promptbox}
The runner defaults to six QA frames and a 16-token action-generation
budget, both configurable. It parses the final action mention, preferring
an explicit action-labelled response when present. A parseable
\texttt{Done} ends the rollout; instance-based success is then computed
from the simulator state. A type-level visibility instruction and an
instance-level success test are therefore distinct parts of the interface.

\subsection{Question and evidence record}
A Rooms-visited schema illustration is:
\begin{promptbox}{Rooms-visited record illustration}
Question: How many different rooms did the robot enter during its run?\\
Options: A: 5 \quad B: 3 \quad C: 4 \quad D: 2\\
Answer: C \quad frame\_refs: [0, 23, 35]
\end{promptbox}
Here, the answer is four rooms while the evidence field contains three
frame references. The prompt and resolved images define the actual input
to the answerer; the record preserves the link between that input and
the trajectory question.

\section{Presence-change evidence selection}
\label{app:presence_change}
The question builder defines disappearance by comparing an object's
last visible position with the final position in the sampled strip. It then
constructs \texttt{frame\_refs} from the first and last visible positions.
For a positive disappearance example, this selection omits the final
observation supporting absence. The gold answer can therefore be correct for
the full strip while unsupported by the cited subset. The sample
in \cref{app:examples} exhibits exactly this mismatch.

The corresponding evidence check re-derives the label from the full
reference sequence and verifies that the selected images establish both
visibility and subsequent absence. Including the later or final frame
supporting disappearance aligns the cited subset with the trajectory-level
answer. This separates label validation from evidence selection.

\subsection{Visibility and evidence support}
Let $V_t(o)$ indicate that object $o$ appears in the visible-object record
at sampled frame $t$, and let $T$ be the final sampled frame. A positive
disappearance example has $V_t(o)=1$ at an earlier frame and $V_T(o)=0$.
This concerns visibility from the camera, not physical removal of the object.
For the training example in \cref{app:examples}, the sequence is
$(1,1,0,0,0,0,0)$ for the bed, while the cited subset is frames $\{0,1\}$.
The full sequence supports disappearance; the cited subset establishes only
the earlier presence.

An evidence-sufficiency check therefore first re-derives the answer
from the ordered symbolic record and then checks that the answerer's selected
observations contain the relevant states in temporal order. Including a
frame with the later absence provides the complementary observation.
An object leaving the field of view within the same room also explains why
this event need not coincide with the generator's room-change-based
\texttt{memory\_required} heuristic.

\section{Oracle input-count shortcut}
\label{app:shortcut}
The joint counts below give the number of cited frames and the gold
room count for Rooms-visited. Their diagonal contains 69 of 116 questions,
yielding 59.5\% for an image-count-only rule. No count exceeds the gold
room count in this table, so input size also supplies a lower bound.

\begin{table}[h]
    \centering
    \small
    \caption{Cited-frame count against gold room count for the 116
    Rooms-visited questions. The distribution never exceeds the diagonal, so the
    frame count is a lower bound on the answer, and equals it in 69 cases
    (bold). Cells are shaded by count.}
    \label{tab:shortcut}
    \begin{tabular}{lrrrr}
        \toprule
        \hdr & \multicolumn{4}{c}{gold room count} \\
        \cmidrule(l){2-5}
        \hdr cited frames & 2 & 3 & 4 & total \\
        \midrule
        1 & \cnt{6}{30} & -- & -- & 6 \\
        2 & \cntb{30}{30} & \cnt{16}{30} & \cnt{1}{30} & 47 \\
        3 & -- & \cntb{18}{30} & \cnt{24}{30} & 42 \\
        4 & -- & -- & \cntb{21}{30} & 21 \\
        \midrule
        total & 36 & 34 & 46 & 116 \\
        \bottomrule
    \end{tabular}
\end{table}

The E2 and E5 scores of 67.2\% and 68.1\% are approximately eight points
above this rule. Subtracting scores from two different predictors does not
decompose the oracle's decisions into visual and non-visual contributions.
The difference can also reflect interactions with choices, count priors,
and prompt cues. A stronger control would use a fixed image count independent
of the answer and compare real images with content-ablated images under
otherwise identical inputs. The exact cited-frame shortcut does not apply
to E0/E0b, which receive no images, or directly to E3/E4, which use their own
observations; other input-count correlations have not been excluded.

\section{Per-stratum navigation detail}
\label{app:strata}
The termination table and full ladder complement the answer scores
with navigation and parsing measurements. The termination categories below
distinguish target acquisition, positional stalling, and target visibility
at termination.

\begin{table}[h]
    \centering
    \small
    \caption{Termination and action-format statistics by stratum.
    Each stratum contains 50 designated episodes. Fmt. denotes action-format
    compliance; Inv. denotes the invalid-action rate. Reached,
    Stuck, and Not vis.\ are termination counts, shaded from white (0) to
    lilac (50).}
    \label{tab:strata_nav}
    \begin{tabular}{llrrrrr}
        \toprule
        \hdr Rung & Stratum & Reached & Stuck & Not vis. & Fmt.\ (\%) & Inv.\ (\%) \\
        \midrule
        \multirow{4}{*}{E1 (random)} & S1 & \cnt{0}{50} & \cnt{0}{50} & \cnt{47}{50} & 54 & 46 \\
         & S2 & \cnt{0}{50} & \cnt{0}{50} & \cnt{48}{50} & 40 & 60 \\
         & S3 & \cnt{0}{50} & \cnt{0}{50} & \cnt{50}{50} & 40 & 60 \\
         & S4 & \cnt{0}{50} & \cnt{0}{50} & \cnt{50}{50} & 48 & 52 \\
        \midrule
        \multirow{4}{*}{E3 (closed-loop)} & S1 & \cnt{0}{50} & \cnt{42}{50} & \cnt{8}{50} & 100 & 0 \\
         & S2 & \cnt{0}{50} & \cnt{45}{50} & \cnt{5}{50} & 100 & 0 \\
         & S3 & \cnt{0}{50} & \cnt{46}{50} & \cnt{4}{50} & 100 & 0 \\
         & S4 & \cnt{0}{50} & \cnt{39}{50} & \cnt{11}{50} & 100 & 0 \\
        \midrule
        \multirow{4}{*}{E5 (oracle replay)} & S1 & \cnt{50}{50} & \cnt{0}{50} & \cnt{0}{50} & 100 & 0 \\
         & S2 & \cnt{49}{50} & \cnt{0}{50} & \cnt{0}{50} & 100 & 0 \\
         & S3 & \cnt{50}{50} & \cnt{0}{50} & \cnt{0}{50} & 100 & 0 \\
         & S4 & \cnt{49}{50} & \cnt{0}{50} & \cnt{0}{50} & 100 & 0 \\
        \bottomrule
    \end{tabular}
\end{table}

\textbf{Route execution across strata.}
The reference replay traverses 6.58\,m in S1 and 22.91\,m in S4, with SPL
0.966 and 0.875, respectively. E3 covers 0.71--1.49\,m across strata and
terminates as stuck in 172 episodes. The random policy instead has a
46--60\% invalid-action rate and 195 recorded target-not-visible outcomes.
Thus action-format errors and limited positional progress produce distinct
execution profiles even when neither policy reaches a target.

\textbf{Stopping behaviour.}
The positional detector increments after an unsuccessful action or at most
0.01\,m of horizontal displacement, and terminates after eight consecutive
such steps. Rotation and camera-pitch changes can therefore collect new views
while advancing the detector. E3's immediate-\texttt{Done} rates are 0.04
in S2 and S3 and 0.12 in S4. These measurements make action interpretation,
stopping, and route extent separate axes for analysing the closed-loop policy.

\section{Full baseline ladder}
\label{app:full_ladder}
\Cref{tab:ladder} brings model identity, navigation, answer coverage, and
answer accuracy into one view. Each column answers a different question:
SR records whether the target instance was reached; SPL measures successful
navigation efficiency; Parse records whether a response could be scored;
and EQA measures answer correctness under the condition's evidence protocol.
Reading the columns together distinguishes an answer-producing system from
an agent that also completes the navigation objective.

\begin{table}[ht]
    \centering\small
    \caption{Interpretation of the principal ladder comparisons. Shared
    components identify the comparison's scope; changed components define
    the intervention that each score comparison represents.}
    \label{tab:comparison_scope}
    \begin{tabular}{>{\raggedright\arraybackslash}p{0.18\linewidth}
                    >{\raggedright\arraybackslash}p{0.32\linewidth}
                    >{\raggedright\arraybackslash}p{0.36\linewidth}}
    \toprule
    \hdr Comparison & Shared components & Changed components \\
    \midrule
    E0b / E0 & Model and answer choices & Question text and target-room label \\
    E3 / E5 & Model and navigation objective & Executed trajectory and QA frame selection \\
    E2 models & Questions and cited-frame protocol & Answering model and its input processing \\
    E3 / E4 & Simulator and instance scorer & Model, local graph input, prompts, and QA trigger \\
    MomaGraph QA configurations & Answering model & Single/multiple frames and QA path \\
    \bottomrule
    \end{tabular}
\end{table}

\textbf{Text-only controls and navigation sanity checks.}
E0b and E0 use no images and have no navigation score. They establish answer
performance available from choices alone or from the question and target-room
label. E1 checks immediate stopping and random action selection. These
policies supply no model-generated answers, so their answer score is n/a;
their zero parse value denotes an inactive answer channel.

\textbf{Model identity and evidence access.}
E3 and E5 use Qwen3-VL-4B, allowing a same-model comparison of self-collected
and reference-trajectory evidence. E5 supplies both route execution and
question-specific frame selection. \rev{E2 uses stored cited frames and
scores 58.5\% with the same model, 55.2\% with MomaGraph-R1, and 55.0\% with
Qwen2.5-VL-7B.}

\begin{table}[h]
    \centering
    \small
    \caption{Full baseline ladder. SR is navigation success rate,
    SPL is success weighted by path length, Parse is answer coverage, and
    EQA is answer accuracy. The designated set contains 200 episodes and
    800 questions. E5 SR and answer metrics use replay-scored episodes
    and questions. E4 accuracy is shown on parsed responses.}
    \label{tab:ladder}
    \scorekey\par\smallskip
    \begin{tabular}{llccccl}
        \toprule
        \hdr Rung & Model & SR & SPL & Parse & EQA (\%) \\
        \midrule
        E0b (options-only)   & Qwen3-VL-4B   & --      & --    & 1.000 & \acc{39.0} \\
        E0 (blind)           & Qwen3-VL-4B   & --      & --    & 0.995 & \acc{45.4} \\
        E1 (immediate-Done)  & --            & 0/200   & 0.000 & 0.00  & n/a \\
        E1 (random action)   & --            & 0/200   & 0.000 & 0.00  & n/a \\
        E3 (closed-loop)     & Qwen3-VL-4B   & 0/200   & 0.000 & 1.000 & \acc{45.5} \\
        E4 (spec.\ closed-loop) & MomaGraph-R1 & 0/200 & 0.000 & 0.16  & n/a (42.2 on parsed) \\
        E2 (frame-oracle)    & MomaGraph-R1  & --      & --    & 1.000 & \acc{55.2} \\
        E2 (frame-oracle, 2nd model) & Qwen2.5-VL-7B & -- & -- & 1.000 & \acc{55.0} \\
        \rev{E2 (frame-oracle, ladder model)} & \rev{Qwen3-VL-4B} & \rev{--} & \rev{--} & \rev{1.000} & \accr{58.5} \\
        E5 (oracle replay)   & Qwen3-VL-4B   & 100\% & 0.924 & 1.000 & \acc{57.4} \\
        \bottomrule
    \end{tabular}
\end{table}
\FloatBarrier

\section{Specialised closed-loop condition}
\label{app:e4}
\subsection{Local scene graph, navigation, and answer trigger}
The E4 adapter implements three stages. At each step, MomaGraph-R1 predicts
a scene graph from the current image, with the first episode question as
context. The graph prompt distinguishes the observed \texttt{scene\_type}
from the instruction-conditioned \texttt{target\_scene}. Its nodes represent
visible objects and its edges describe spatial and functional relations.
The action stage receives the current image, a formatted local graph summary,
and the target-room and target-type instruction. The summary lists the
predicted room, objects, and relations for that step; the action prompt does
not accumulate earlier scene graphs.

The answer stage executes only after the model emits \texttt{Done}, using
the terminal image. When the rollout terminates through the stuck detector
or budget instead, the adapter writes a null answer with
\texttt{parse\_ok: false}. This control flow distinguishes answer availability
from the model's ability to answer when invoked. The template is:
\begin{promptbox}{E4 terminal-image answer prompt}
You have navigated to the \{target\_room\} and are looking at the
\{target\_object\}.\\[3pt]
Answer the following question.\\
Question: \{question\}\\
Options:\\
\{options\}\\[3pt]
Respond with ONLY the answer letter (e.g. 'A').
\end{promptbox}
The first sentence is a template instruction. The success scorer separately
tests target-instance distance and image coverage, so an emitted
\texttt{Done} can trigger QA even when the navigation objective is not met.
The adapter defaults to 1,024 tokens for scene-graph prediction, 16 for the
action, and 10 for the answer. Its GPU loader uses \texttt{float16}.

\subsection{Navigation and answer measurements}
Over 200 episodes, E4 reaches zero targets (SPL 0.000) with 100\%
action-format compliance, 0\% invalid actions, mean distance reduction
0.17\,m, and mean executed path 0.46\,m. It terminates as stuck in 168
episodes and target-not-visible in 32, with an immediate-\texttt{Done} rate
of 0.06. It parses 128/800 answers (16\%), of which 54 are correct
(42.2\%); 54/800 gives the rounded unconditional score of 6.8\%.
Unconditional family accuracies range from 0.044 to 0.098.
The answer log separates model responses from parsed letters, supporting
analysis of evidence selection, generation, and parsing at the question level.

\textbf{Comparison with frame-based answering.}
MomaGraph-R1 scores 44.1\% in its single-frame QA configuration and 55.2\%
with multi-frame oracle evidence. E4's 42.2\% parsed-answer score lies near
the single-frame result, but the question coverage and evidence protocols
differ. Reporting all three quantities makes the role of answer availability
visible while preserving the evidence that the model answers more accurately
under the multi-frame configuration.

\textbf{Relation to the generic closed-loop baseline.}
Both E3 and E4 produce fully parsed action names and reach no target under
the evaluated harness. Their mean paths, 1.03\,m and 0.46\,m, and stuck
counts, 172 and 168, describe a common pattern of limited displacement.
This comparison includes both a generic VLM and a model specialised for
embodied scene-graph prediction, with their respective action and QA protocols.
The result motivates testing persistent
state and observation-aware stopping within an otherwise matched action
interface; it does not attribute the shared outcome to a particular memory
representation.

\section{Data and execution examples}
\label{app:examples}

\subsection{A trajectory-linked question record}
The training sample \texttt{ep\_0001} is an S1 Bedroom episode
with target instance \texttt{Painting|2|5}, seven frames, and zero doorway
teleport hops. The following fields are copied from its second question:
\begin{promptbox}{Example question record}
question\_id: ep\_0001\_q2\\
subcategory: Presence-change\\
question: Was a bed visible at some point but then gone from view later?\\
options: [A: No, B: Yes]\\
answer: B\\
frame\_refs: [0, 1]\\
memory\_required: false
\end{promptbox}
The frame records and corresponding images support this sequence:
\begin{center}\small
\begin{tabular}{rlll}
\toprule
\hdr Frame & Recorded action & Room & Bed in visible-object log \\
\midrule
0 & Spawn & Bedroom & Yes \\
1 & RotateLeft & Bedroom & Yes \\
2 & RotateLeft & Bedroom & No \\
3 & MoveAhead & Bedroom & No \\
4 & RotateLeft & Bedroom & No \\
5 & LookUp & Bedroom & No \\
6 & Done & Bedroom & No \\
\bottomrule
\end{tabular}
\end{center}

\begin{figure}[h]
    \centering
    \includegraphics[width=0.29\linewidth]{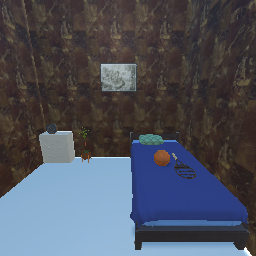}\hfill
    \includegraphics[width=0.29\linewidth]{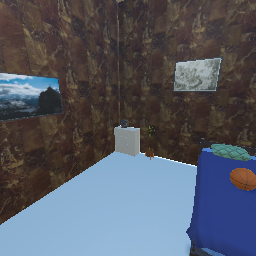}\hfill
    \includegraphics[width=0.29\linewidth]{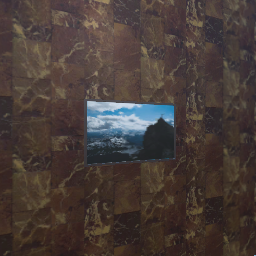}
    \caption{Training sample: frame 0 (left) and frame 1 (middle)
    are cited; frame 6 (right) is not. The bed appears in the cited evidence
    and is absent from the final view. This training example illustrates how the trajectory, question,
    and selected evidence are linked.}
    \label{fig:sample}
\end{figure}
The full trajectory supports answer B, but the cited pair does not show the
later absence. The \texttt{memory\_required} flag is false because its
heuristic requires a room change after the cited evidence; camera motion can
still move an object out of view within one room. Thus both evidence
references and memory flags need interpretation at the instance level.

\subsection{A paired execution example}
An S3 test episode from a pilot run of the same evaluation pipeline
illustrates the paired navigation and answer records. Episode identifiers
are scoped by split and run, so this \texttt{ep\_0001} is distinct from the
training sample above. The two conditions record:
\begin{center}\small
\begin{tabular}{lrr}
\toprule
\hdr Recorded field & E3 closed loop & E5 oracle replay \\
\midrule
Actions & 8 & 30 \\
Invalid actions & 0 & 0 \\
Executed path (m) & 0.00 & 10.75 \\
Reached target & No & Yes \\
Termination & stuck & reached\_target \\
Correct / scored questions & 2/2 & 2/2 \\
\bottomrule
\end{tabular}
\end{center}
For question \texttt{ep\_0001\_q1}, the E5 prompt asks
``Was a toilet visible in the kitchen?'' with choices A: Partially visible,
B: No, C: Cannot determine, and D: Yes, citing frame 18. Both conditions
record \texttt{raw: "B"}, \texttt{parsed: "B"}, and
\texttt{parse\_ok: true}, so identical answer scores coexist with different
navigation outcomes.

\subsection{Paired navigation and answer records}
The following selected fields are copied from the example's navigation
and prediction records. The shared question identifier connects the answer
record to its episode within the run; the run and split identify the
surrounding dataset context.
\begin{promptbox}{Closed-loop navigation log: selected fields}
episode\_id: ep\_0001\\
reached\_target: false\\
path\_length\_replayed\_m: 0.0\\
final\_dist\_to\_target\_m: 4.5744\\
n\_actions: 8\\
n\_invalid\_actions: 0\\
termination: stuck
\end{promptbox}
\begin{promptbox}{Oracle-replay navigation log: selected fields}
episode\_id: ep\_0001\\
reached\_target: true\\
path\_length\_replayed\_m: 10.75\\
path\_length\_optimal\_m: 8.08747792727554\\
final\_dist\_to\_target\_m: 1.2511\\
n\_actions: 30\\
n\_invalid\_actions: 0\\
termination: reached\_target
\end{promptbox}
\begin{promptbox}{Answer log: identical in the two conditions}
question\_id: ep\_0001\_q1\\
raw: B\\
parsed: B\\
parse\_ok: true\\
pred: B
\end{promptbox}
The corresponding per-episode records give EQA accuracy 1.0 with two
correct, scored, and parsed questions in each condition. Replay SPL is
0.7523, consistent with $8.0875/10.75$ for this successful episode; closed-loop
SPL is zero because its target was not reached. The navigation and answer
records jointly expose a case that answer accuracy alone would treat as a tie.

\subsection{Reading a trajectory-linked evaluation record}
An episode supplies the spawn pose, target-instance record, ordered reference
actions, and observed frames. A question then supplies its family, choices,
gold answer, and indices into that reference sequence. The evidence condition
determines which observations reach the answerer, with no images for E0b/E0,
stored cited images for E2, sampled rollout images for E3, the terminal
image after \texttt{Done} for E4, and replayed
cited images for E5. The final prediction record preserves both raw text and
the parsed answer. Navigation metrics and answer metrics can then be joined
by run, split, episode, and question identifiers.

\section{Execution and reproducibility}
\label{app:repro}
The evaluation stack combines simulator rendering, model inference, and
structured scoring, and the run record preserves each stage alongside the
final metrics. Every rollout logs controller initialisation, spawn
restoration, image count, model-call count, and answer parsing separately.
The 200-episode test split, with its 800 questions, rendered frames, and
per-episode generation settings, is released at
\url{https://huggingface.co/datasets/Divyansh0108/Vroom-To-Room-200}.

\subsection{Coverage and metric denominators}
For a designated question set $\mathcal Q$, let $P_q$ indicate a parsed
response and let $C_q$ indicate a correct parsed response. Define
\begin{align}
 r_{\mathrm{parse}} &= \frac{\sum_{q\in\mathcal Q}P_q}{|\mathcal Q|}, &
 a_{\mathrm{all}} &= \frac{\sum_{q\in\mathcal Q}C_q}{|\mathcal Q|}, &
 a_{\mathrm{parsed}} &= \frac{\sum_{q\in\mathcal Q}C_q}{\sum_{q\in\mathcal Q}P_q}.
\end{align}
For nonzero parse coverage,
$a_{\mathrm{all}}=r_{\mathrm{parse}}a_{\mathrm{parsed}}$.
E4 illustrates the identity, where $128/800$ parse coverage and $54/128$
conditional accuracy yield $54/800$, or 6.8\% rounded, over the designated
questions.
When no responses are parsed, the conditional score is undefined.

Execution coverage is a separate denominator. Accuracy on an executed
subset is always accompanied by the fraction of designated questions it
covers, and the designated, executed, scored, and parsed sets are recorded
so that both the end-to-end and the conditional score can be recomputed.

\subsection{Object vocabulary and observation normalization}
The observation vocabulary contains 104 canonical object types.
Normalization maps observed names to these types, and duplicate type
mentions are merged before per-room truncation. A memory-conditioned model therefore depends on the
vocabulary, normalization, deduplication, and truncation settings as well as
its weights, so these transformations are versioned together with the
observation cache.

\subsection{Run manifest and record linkage}
A run manifest associates the dataset version and split-aware house
identity with episode and question identifiers, simulator settings, model
revision, decoder settings, and the evidence policy of the condition. The
designated episode set is stored separately from execution status.
Navigation records capture path length, actions, target success, and
termination; prediction records capture raw text, parsed letters, and parse
status (\cref{tab:log_fields}).

\begin{table}[ht]
    \centering\small
    \caption{Fields of the navigation and prediction records. Paths and
    distances are in metres; counts are integers. Episode and question
    identifiers are interpreted within their run and split.}
    \label{tab:log_fields}
    \begin{tabular}{>{\raggedright\arraybackslash}p{0.46\linewidth}
                    >{\raggedright\arraybackslash}p{0.44\linewidth}}
    \toprule
    \hdr Field & Interpretation \\
    \midrule
    \texttt{episode\_id} & Episode key for navigation and per-episode metrics \\
    \texttt{question\_id} & Question key for raw and parsed predictions \\
    \texttt{reached\_target} & Outcome of the target-instance success check \\
    \texttt{path\_length\_replayed\_m} & Executed path, including in the closed-loop log \\
    \texttt{path\_length\_optimal\_m} & Reference path distance recorded for replay \\
    \texttt{final\_dist\_to\_target\_m} & Distance to the target at termination \\
    \texttt{n\_actions}, \texttt{n\_invalid\_actions} & Executed-action and invalid-action counters \\
    \texttt{termination} & Recorded stopping outcome \\
    \texttt{raw}, \texttt{parsed}, \texttt{parse\_ok} & Model response, extracted letter, and parsing status \\
    \texttt{n\_correct}, \texttt{n\_total}, \texttt{n\_parsed} & Per-episode answer counts used in aggregation \\
    \bottomrule
    \end{tabular}
\end{table}
\end{document}